\pdfoutput=1

\documentclass[11pt]{article}

\usepackage[]{acl}

\usepackage[]{acl}
\usepackage{inconsolata}

\usepackage{hyperref}
\usepackage{url}
\usepackage{times}
\usepackage{latexsym}
\usepackage{url}
\usepackage{booktabs}
\usepackage{multirow}
\usepackage{graphicx}
\usepackage{pifont}
\usepackage{pythonhighlight}

\usepackage{times}
\usepackage{latexsym}

\usepackage[T1]{fontenc}

\usepackage[utf8]{inputenc}

\usepackage{microtype}

\usepackage{inconsolata}

\usepackage{graphicx}

\title{Walia-LLM: Enhancing Amharic-LLaMA by Integrating Task-Specific and Generative Datasets}

\author{\normalsize Israel Abebe Azime $^{1}$,  Atnafu Lambebo Tonja $^{2,3}$,  Tadesse Destaw Belay $^{2}$, Mitiku Yohannes Fuge $^{4}$,  \\ 
\textbf{\normalsize Aman Kassahun Wassie$^{4}$,  
Eyasu Shiferaw Jada, Yonas Chanie$^{5}$, Walelign Tewabe Sewunetie $^{6}$, } \\
\textbf{\normalsize Seid Muhie Yimam $^{7}$} \\\\
\footnotesize
$^\forall$ Masakhane NLP, $^\forall$ Ethio NLP, $^1$ Saarland University, Germany, , $^2$ Instituto Politécnico Nacional, Mexico,$^3$ Lelapa AI,$^4$  AIMS
\\
 \footnotesize
 $^5$ Carnegie Mellon University, $^6$ Debre Markos University, $^7$Universität Hamburg, Germany
}

\begin{document}
\maketitle
\begin{abstract}
Large language models (LLMs) have received a lot of attention in natural language processing (NLP) research because of their exceptional performance in understanding and generating human languages. However, low-resource languages are left behind due to the unavailability of resources. In this work, we focus on enhancing the \textsc{LLaMA-2}-Amharic model by integrating task-specific and generative datasets to improve language model performance for Amharic. We compile an Amharic instruction fine-tuning dataset and fine-tuned \textsc{LLaMA-2}-Amharic model. The fine-tuned model shows promising results in different NLP tasks. We also explore the effectiveness of translated instruction datasets compared to the dataset we created. Our dataset creation pipeline, along with instruction datasets, trained models, and evaluation outputs, is made publicly available to encourage research in language-specific models.
\footnote{For data generation pipeline, see \url{https://github.com/EthioNLP/afri-sft-data}. For models and datasets, refer to \url{https://huggingface.co/EthioNLP}. }
\end{abstract}

\section{Introduction}

Large language models (LLMs) such as GPT series \citep{brown2020language}, \textsc{Llama-2} \citep{touvron2023LLaMA}, Phi2 \citep{javaheripi2023phi}, Mistral \citep{jiang2023mistral}, Mixtral \citep{jiang2024mixtral}, PaLM \citep{chowdhery2023palm}, Gemini \citep{team2023gemini},
BLOOM \citep{workshop2022bloom}, have exhibited exceptional performance in understanding and generating human language, showcasing a range of capabilities from basic linguistic comprehension to complex text generation. 
 
\textsc{Llama-2} \citep{touvron2023LLaMA}, a family of pre-trained and fine-tuned large language models (LLMs), demonstrated impressive performance across multiple tasks, particularly in dialogue-based interactions. Regardless of these achievements, \textsc{Llama-2} pre-training supports a small number of languages, which does not include low-resource languages like Amharic.   This makes adapting LLMs to low-resource languages that are not included a significant challenge.

Adopting these LLMs to local languages requires the preparation of a quality instruction dataset. Amharic is one of the Semitic languages under the Afroasiatic language family spoken in Ethiopia with more than 57M speakers. There are numerous task-specific datasets for Amharic \citep{tonja-etal-2023-natural} compared to other Ethiopian languages. This paper focuses on enhancing the \textsc{Llama-2}-Amharic \citep{andersland2024amharic} model with quality datasets that are created by converting existing datasets in English into instruction-based Amharic datasets. Furthermore, we create new instruction datasets following the approach by \citet{wei2021finetuned}.

\textsc{Llama-2}-Amharic model by \citet{andersland2024amharic} was created by pre-training \textsc{Llama-2} 7B model using open-source Amharic and translated corpus. After performing vocabulary expansion and pre-training, \citet{andersland2024amharic} fine-tuned the created model by translating  English instruction datasets into Amharic using commercial translation tools. In our research, we aim to improve the performance of the Amharic \textsc{LLaMA} model by integrating task-specific and generative datasets, as shown in Table \ref{tab:newdata}. 
The contributions of this paper are as follows: 
\begin{itemize}
    \item Creating Amharic instruction fine-tuning data from existing NLP task-specific and generative datasets. 
    \item Evaluating new and existing models' performance. 
    \item Exploring the effect of carefully curated datasets by combining them with machine-translated instruction datasets.
    \item Exploring the effect of instructions on the model's performance by introducing code-mixing instructions. 
    \item Open-sourcing our dataset creation pipeline, instruction datasets, trained models, and evaluation outputs to promote language-specific studies on these models.
\end{itemize}

\section{Related Work}
The introduction of open-source LLMs like \textsc{Llama-2} \citep{touvron2023LLaMA} enabled the creation of several language models that focus on specific applications. This application gives more capabilities for these LLMs by teaching them to use tools \citep{schick2023toolformer}, write code \citep{roziere2023code}, understand videos \citep{damonlpsg2023videollama}, or work for different languages \citep{cui2023efficient}. To achieve remarkable understanding and generation abilities, LLMs require large training data and huge compute resources \citep{hoffmann2022training}.

The work by \citet{dong2023abilities} explores how LLMs' generation, natural language understanding, and problem-solving abilities relate to the data they are trained on and its composition. This work suggests that the amount of composition data is more important for these abilities to show in a low-resource scenario.

Using self-instructed fine-tuning, the work by  \citet{wei2021finetuned,alpaca,cui2023efficient} showed a new approach to align the generation outputs of the generative models through the application of NLP tasks. These tasks are structured around natural language instruction templates, providing a novel means to guide the model's generation process toward better adherence to task-specific requirements. \textsc{LLaMA}-Adapter \citep{zhang2023llama} also shows that it is possible to reduce the fine-tuning time of \textsc{LLaMA}-7B by introducing lightweight adapters on top of the model.

Acquiring and preparing a dataset for instruction fine-tuning presents a significant challenge due to the extensive labor and resources required. There are several ways of acquiring instruction data, including manual dataset creation, using generative models \citep{wang2022self, alpaca}, or using machine translation instruction data for training LLMs for specific languages \citep{cui2023efficient}.

Fine-tuning LLMs such as \textsc{Llama-2} for specific tasks is an area of exploration as well. Advanced language model-based translator (ALMA) \citep{xu2023paradigm} outperformed state-of-the-art (SOTA) no language left behind (NLLB) \citep{nllb2022} model MT task. They worked on fine-tuning monolingual data and subsequent fine-tuning with parallel data. Apart from \textsc{Llama-2}, \cite{moslem2023fine} worked on Mistral 7B fine-tuning for medical domain machine translation, where they showed improvement in Spanish to English translation from baseline performance. 

After the \textsc{Llama-2} was released, researchers successfully adapted the model for other languages. The work by \citet{cui2023efficient} involved creating a unique tokenizer for Chinese, extending the pre-training phase, and then fine-tuning the model. This work incorporates secondary pre-training using Chinese data and fine-tunes the model with Chinese instruction datasets. The result shows a significant enhancement of the model’s ability to comprehend and execute instructions. 

To the same approach of the work by \citet{cui2023efficient}, \textsc{Llama-2} was also adopted for Amharic \citep{andersland2024amharic}. During pre-training, \citet{andersland2024amharic} used an open-source Amharic corpus with some translated corpus from English, and for fine-tuning, available English instruction datasets were translated to Amharic using the Google Translate API and SeamlessM4T. Following the increase of the \textsc{LLaMA} vocabulary size from 32k to 51k and subsequent pre-training with a large Amharic text corpus, they conducted supervised instruction fine-tuning using machine-translated datasets. Then, they evaluated their model using the MMLU \citep{hendrycks2020measuring} multiple-choice English dataset by translating it into Amharic. The model is available without original Amharic evaluations because no instruction-based datasets exist for Amharic. 

\begin{table*}[!ht]
\footnotesize
\centering
\scriptsize
\begin{tabular}{lrrrrrrrr}
\toprule
\textbf{Data Source} & \multicolumn{3}{c}{\textbf{Source Data}} & \multirow{2}{*}{\textbf{Is new}} & \multirow{2}{*}{\textbf{\# Templates}} & \multicolumn{3}{c}{\textbf{Generated Data}} \\ 
  & \textbf{train} & \textbf{val} &\textbf{test} &  &   & \textbf{train} &\textbf{val} &\textbf{test}  \\
\midrule
Amharic QA & 1723 & 595 & 299 & NO & 14 & 10000 & 595 & 299  \\
MasakhaNews  & 11522 & 188 & 376 & NO & 11 & 7866 & 205 & 376  \\
MT (amh-eng) & 497739 & 1012 & 1012 & NO & 10 & 10000 & 997 & 1012 \\
MT (eng-amh)  & 497739 & 1012 & 1012 & NO & 10 & 10000 & 997 & 1012  \\
Summarization  & 5761 & 719 & 719 & NO &  9 & 10000 & 719 & 719  \\
Text Expansion  & 5761 & 719 & 719 & NO &  9 & 10000 & 719 & 719  \\
Sentiment Analysis (AfriSenti) & 5984  & 1497 & 1999 & NO &  7 & 10000 & 1728 & 1999 \\
NER & 1750  & 500 & 250 & NO & 9 & 10000 & 500 & 250    \\
News Title Generation & - & - & - & Yes & 10 & 10000 & 5078 & 5078   \\
Poem Generation & - & - & - & Yes & 3 & 3885 & 69 & 70   \\
Poem Completion & - & - & - &  Yes& 7 & 3885 & 69 & 70  \\
Religious Lyrics  Generation & - & -  & -  & Yes & 3 & 4929 & 188 & 206  \\
Religious Lyrics  Completion & - & - & -  & Yes & 4 & 10000 & 1497 & 1728  \\
Story generation & - & - & - & Yes & 10& 1665 & 24 & 25  \\
Spelling Correction & - & -  & - & Yes (modified) & 9 & 10000 & 1438 & 	1438     \\
Non religious music Lyrics Generation  & - & -  & - & Yes & 4 & 148 & 5 & 5   \\
Non religious music Lyrics Completion  & - & -  & - & Yes & 7 & 259 & 5 & 5   \\
\midrule
Total &  &   &  &  & & 122,637  & 14,911 & 15,011   \\
\bottomrule
\end{tabular}
\caption{\textbf{Dataset} used for preparing instruction fine-tuning data. \textbf{Is new} = new custom dataset. Details of each data source are explained in Section \ref{sec:data}. Figure \ref{fig:datapipeline} shows how a dataset is converted to instruction data using our Data processing Pipeline. }
\label{tab:newdata}
\end{table*}

\section{Dataset preparation}
\label{sec:data}
\begin{figure*}[t!]
    \centering
\includegraphics[width=14cm]{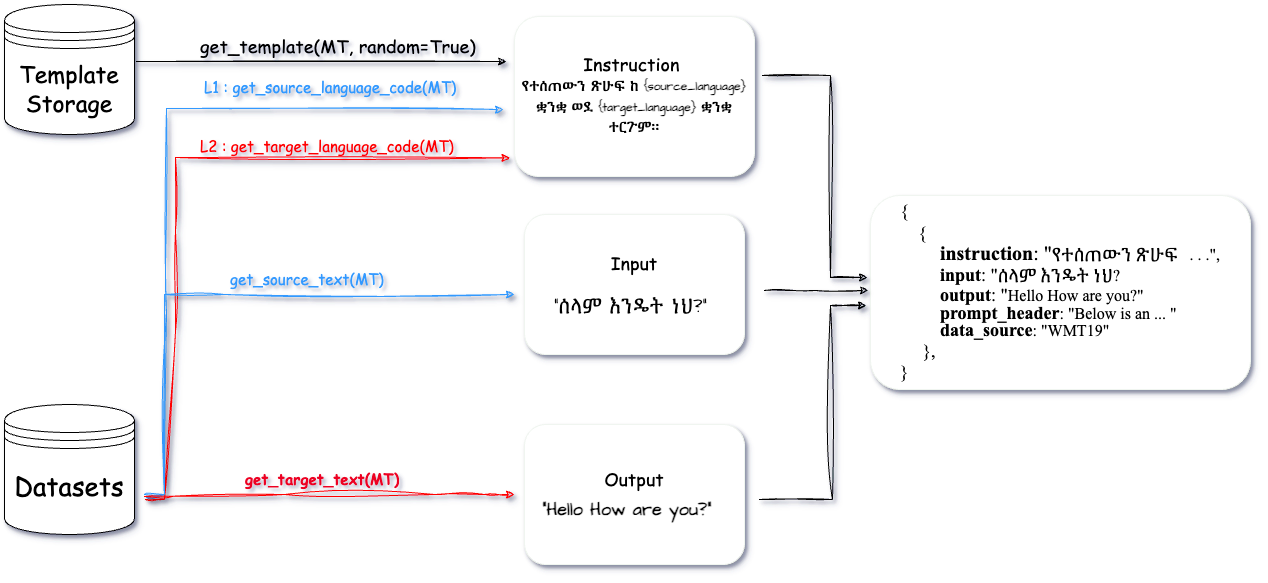}
    \caption{Data processing Pipeline. The pipeline creates instruction data from existing task datasets, and from generative datasets, we collected. All instructions, input, and output are in Amharic except for the MT case, as shown in the picture. The data source will not be used during training.}
    \label{fig:datapipeline}
\end{figure*}

In this work, we have converted existing NLP task-specific datasets, like sentiment analysis and machine translation, into instruction datasets. We created an instruction template for each task and developed a data creation pipeline (Figure \ref{fig:datapipeline}) that merges each template instruction with appropriate data from a pre-existing dataset. This pipeline helps us to create instruction datasets from pre-existing NLP task datasets. For the new NLP task, we focused on collecting a new dataset that can be converted into instruction data. We also created new datasets by tweaking existing datasets. Finally, we included an instruction-tuning dataset converted into Amharic language using machine translation systems. Table \ref{tab:newdata} shows a detailed distribution of instruction task data.

\subsection{Instruction dataset from existing datasets}\label{data-task-based}
We have used several existing datasets to create an instruction dataset from an existing one. The production of these datasets includes web scraping, human labeling, and verification. By collecting and using this dataset for instruction, we ensure the quality of our instruction dataset. The other benefit of working with these datasets is that we ensure similar prompts across all our models for testing, which eliminates prompt-related performance variance that is usually reported while evaluating the performance of this dataset in generative LLMs. 

For sentiment analysis data, we used \textit{AfriSenti} \citep{muhammad2023afrisenti}, a sentiment analysis benchmark dataset for 14 African languages where Amharic is among the ones. The dataset is labeled with three sentiment classes: positive, negative, and neutral. The number of train, test, and val sets are shown in Table \ref{tab:newdata}. We used the Amharic version of the classes for the test cases, and tests were done to check if the model gives one of the sentiment classes during generation.

We also worked with \textit{MasakhaNews} \citep{adelani2023masakhanews}, a benchmark dataset for news topic classification covering 16 languages widely spoken in Africa. It provides an evaluation of baseline models from classical machine learning models and fine-tunes several language models. 

To test if our model has the ability to identify names from sentences, we modified \textit{MasakhaNER} \citep{adelani2021masakhaner}, which is a dataset for named entity recognition (NER) in ten African languages. We created questions to extract only personal names for this work, and we plan to include more in our future works.

\textit{AmharicQA} \citep{abedissa2023amqa} is a publicly available Amharic open-ended question-answering dataset. It is a crowdsourced 2,628 question-answer pairs over 378 Wikipedia articles. These question-answer pairs are supplemented with context that the language model can use to answer the questions. We have also used this dataset to evaluate our models by converting it into an instruction dataset.

For tasks like Amharic text summarization, we used \textit{XL-Sum} \citep{hasan2021xl}, a comprehensive and diverse dataset comprising 1M annotated article-summary pairs. The dataset covers 44 languages, ranging from low to high-resource ones. We utilized the Amharic portion of the dataset in two ways. First, we took the text and prepared an instructional dataset to test our model's ability to summarize the text. We also created a text expansion task where our model takes the shorter sentence and produces a detailed explanation of the text, the inverse of the text summarization task.

Finally, we used the dataset by \citet{barrault-etal-2019-findings, nllb2022} to prepare training, validation, and testing for machine translation. Our training dataset is from  WMT19 \citep{barrault-etal-2019-findings}, and validation and testing are from \cite{nllb2022}.

The \textit{Amharic spell correction} dataset is designed to assess the effectiveness of models in correcting Amharic spelling errors, covering common misspellings to advance NLP tools for the language. We leveraged Amharic BBC news texts from XL-Sum \citep{hasan2021xl} for this task. We also leveraged the text augmentation library nlpaug \citep{ma2019nlpaug}. We introduced some random character augmentations, including \textit{insertion}, \textit{substitution}, \textit{swapping}, \textit{deletion}, and \textit{word cropping}. This augmentation is done randomly and applied to part of the dataset.

After preparing each dataset, we found that the machine translation dataset we have was significantly larger than the other tasks, so we set a maximum threshold of 10k instructions randomly for the training split of each dataset. For validation and testing, we only used one template per task, and we did not expand the data sizes. More dataset examples and explanations are found in the Appendix \ref{sec:appendix2}.

\subsection{New Custom Datasets}

Most of the task datasets we prepared in Section \ref{data-task-based} did not focus on generation tasks. Generation tasks are less explored for low-resource languages like Amharic, so we created original datasets collected from publicly available sources.


In Amharic, music, stories, and poems represent fascinating cultural artifacts. We have created three new datasets to facilitate the training and evaluation of models' capabilities in processing these tasks. The first track we considered is \textit{religious music lyrics generation}. We included several types of music lyrics in this dataset. We collected the above 2k Amharic spiritual song lyrics from WikiMezmur\footnote{\url{https://wikimezmur.org/am}}. Despite the popularity of non-religious music in Ethiopia, finding a freely available source to include this in our data was difficult; hence, our non-religious music data was smaller than the others. To expand this dataset, we split the lyrics into verses and created a new completion task where the input is the first verse and the output is the remaining whole verse.

To understand the story generation abilities of different models, we created a dataset for \textit{Ethiopian folktales}: We collected several Ethiopian folktales from EthopianFolkTales\footnote{\url{https://www.ethiopianfolktales.com/am}}. These stories are collected from all Ethiopian regions. Given that the dataset comprises traditional Ethiopian stories, there is no copyright restriction on them, and our usage is only for research purposes. We also collected \textit{Amharic poem} from several public telegram channels. 

For \textit{news title generation}, we collected 50k news title and body pairs from different Amharic news sources such as BBC News\footnote{\url{https://www.bbc.com/amharic}}, Deutsche Welle (DW) news\footnote{\url{https://www.dw.com/am}}, Sheger FM\footnote{\url{https://www.shegerfm.com/}}, Addis Admass Newspaper\footnote{\url{https://www.addisadmassnews.com/}}, and VOA Amharic\footnote{\url{https://amharic.voanews.com}}. To save GPT-4 credits, we did our testing only on the first 1300 items of this data.

\subsection{Translated instruction fine-tuning dataset}

During \textsc{Llama} model self-instructed
fine-tuning \citep{touvron2023LLaMA}, instruction datasets like Alpaca \citep{alpaca} and dolly \citep{DatabricksBlog2023DollyV2} have been widely used. In the work \citet{andersland2024amharic}, machine translation systems were used to translate these datasets into Amharic instruction fine-tuning data. This method is used by most papers that try to adopt \textsc{Llama} models for their language, like \cite{cui2023efficient}. For Amharic versions of alpaca and dolly datasets, we used datasets used by \textsc{Llama-2}-Amharic \citep{ andersland2024amharic} training. We explored the effect of training a model by using only our relatively clean and human-verified data alone and in combination with this machine translation data.

\begin{figure*}[ht]
    \centering
\includegraphics[width=16cm]{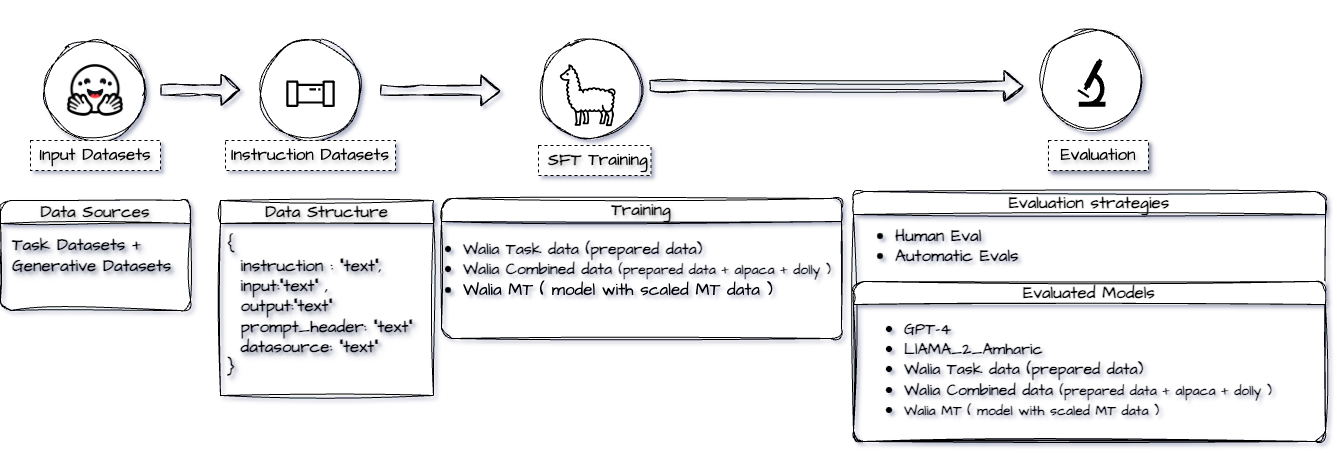}
    \caption{Full training pipeline that summarizes the work done. }
    \label{fig:trainingpipeline}
\end{figure*}
\section{Experiments}

We followed Chinese \textsc{LLaMA} \citep{cui2023efficient} experiments to perform supervised fine-tuning (SFT) on our dataset using different types of the dataset we created. Figure \ref{fig:trainingpipeline} shows the full training pipeline that summarizes the overall experiment steps we followed. We used codes available on the Chinese-\textsc{LLaMA}-Alpaca\footnote{\url{https://github.com/ymcui/Chinese-LLaMA-Alpaca}} repository. We used 4 A100 GPUs with the default parameters in the repository. All training is done for three epochs. All models and evaluation codes will be available in our repository. For the MT task, we also worked on M2M100 \citep{fan2021beyond} and NLLB \citep{nllb2022} models. 

During the evaluation of the models, we used \texttt{gpt-4-0613} for GPT-4. For our \textsc{LLaMA}-based models, we used fixed generation parameters across the models. We also evaluated various models that purportedly support this language but excluded them due to their inability to perform the required tasks. This is further discussed in Appendix \ref{sec:appendix}.

Our main experiment includes:
\begin{itemize}
    \item Evaluating existing models on our dataset.
    
   \item Fine-tuning the model using a dataset detailed in Table \ref{tab:newdata}, referred to as \textbf{Walia (task data)}. Unlike the approach taken in the \textsc{Llama-2}-Amharic model \citep{andersland2024amharic}, this experiment did not incorporate machine-translated instructional data.
    
    \item Fine-tuning the model using \textbf{Walia (combined data)}, which consists of our prepared dataset along with the machine-translated instructional data previously utilized in the \textsc{Llama-2}-Amharic model \citep{andersland2024amharic}.

    \item Fine-tuning \textbf{Walia MT}, the MT model we trained to perform the machine translation task in our dataset. In this experiment, we used only MT datasets from Table \ref{tab:newdata} and scaled them to 200k data rather than 20k as shown in the table.
    
     \item Exploring the effect of prompts in existing and available models for Amharic tasks. Additionally, how code mixing affects the performance of models. This include experiments discussed in \ref{sec:prompt}.
    
\end{itemize}
\subsection{Datasets}
The first set of experiments we conducted involved evaluating the base \textsc{Llama-2}-Amharic model \citep{wei2021finetuned} on our custom test set, which was created from different NLP task datasets. This will provide us with a baseline performance for Amharic tasks. The next set of experiments used different NLP task datasets that were converted into an instruction dataset by our data generation pipeline. We used \textsc{Llama-2}-Amharic model \citep{andersland2024amharic}, which is pre-trained using the \textsc{Llama-2} model for the Amharic language and performed supervised instruction fine-tuning on the task datasets. This ensures our model only has access to quality datasets that were adopted from verified NLP tasks. Finally, we combined our instruction dataset with the machine-translated instruction datasets. In the different datasets above, we have capped our training dataset to a maximum of 10k data from individual tasks, as shown in Table \ref{tab:newdata}. We kept fixed instruction and data frequency in our validation and test set to avoid any performance variation because of instruction differences.
For machine translation experiments, we created additional data that contains 200k data points from \citet{barrault-etal-2019-findings} and \citet{nllb2022}. 

\subsection{Evaluation Metrics}\label{Evaluation}
For selected NLP tasks in this paper, we used different evaluation metrics. For \textit{sentiment analysis} and \textit{news classification} tasks, we have used the weighted f1 score. For these classification tasks, we also keep track of the number of times the model returns output that cannot be classified as one of the classes. 

For \textit{generation tasks}, we used Rogue \citep{lin-2004-rouge} scores. We used Rogue scores to evaluate \textit{xlm-summarization}, \textit{reverse summarization}, and \textit{AmharicQA} tasks. We reported Rogue1, Rogue2, and RugueL metrics for generation tasks, but we heavily rely on RogueL for analysis since it focuses on the longest common subsequence rather than n-grams. We observed that most of our generation outputs do not share common n-grams when n is greater than 2, and the generations from systems like GPT-4 tend to be longer where the n-gram comparison methods express the results less. Additionally, we used word-based evaluation metrics, SacreBLEU \citep{post2018call} and character-based evaluation metrics, chrF++ \citep{popovic2017chrf++} automatic evaluation metrics for MT tasks. We added character-based metrics (chrf++) because, for low-resource languages with complex morphologies, chrF++ offers a more robust and adaptable metric compared to word-based metrics like SacreBLEU. 

Finally, we performed human evaluation for generative tasks such as music, poetry, and story generation. We sampled 120 individual items and conducted blind reviews using three people for each question. We created a rating system from 1 to 5 with detailed instructions and reported the average rating per task and model. 

We did several evaluations for some tasks that were hard to evaluate, e.g., we used accuracy and SacreBLEU scores as evaluation metrics for AmharicQA following the suggestion by \citet{abedissa2023amqa, lee-etal-2021-kpqa}. For tasks that require specific text output, we performed character normalization and text cleaning on the outputs before evaluation to avoid analysis because of typos and formatting issues.

In addition to the evaluation methods mentioned above, we explored the possibility of using GPT-4 for evaluation purposes, following the work from the Chinese \textsc{LLaMA} \citep{cui2023efficient}. Our assessment covered various generation tasks, showing that GPT-4 performs well in most areas. However, it shows inconsistency in scoring due to differences in the rating scale it assigns during each run. In addition, it struggles with evaluating poem and music generation tasks, as it does not fully understand Amharic poetic structure. Additionally, it encounters some challenges in evaluating machine translation, often missing grammatical details in Amharic sentences. Despite these limitations, GPT-4 has the potential for evaluating tasks if it is coupled with manual checks to ensure consistency. We expect similar difficulties in other low-resource languages based on our preliminary findings. While we did not include GPT-4 scores in our current reports due to time and cost constraints, we plan to include them in future research.
\begin{figure}[!ht]%
    \centering
    \qquad{{\includegraphics[width=7cm]{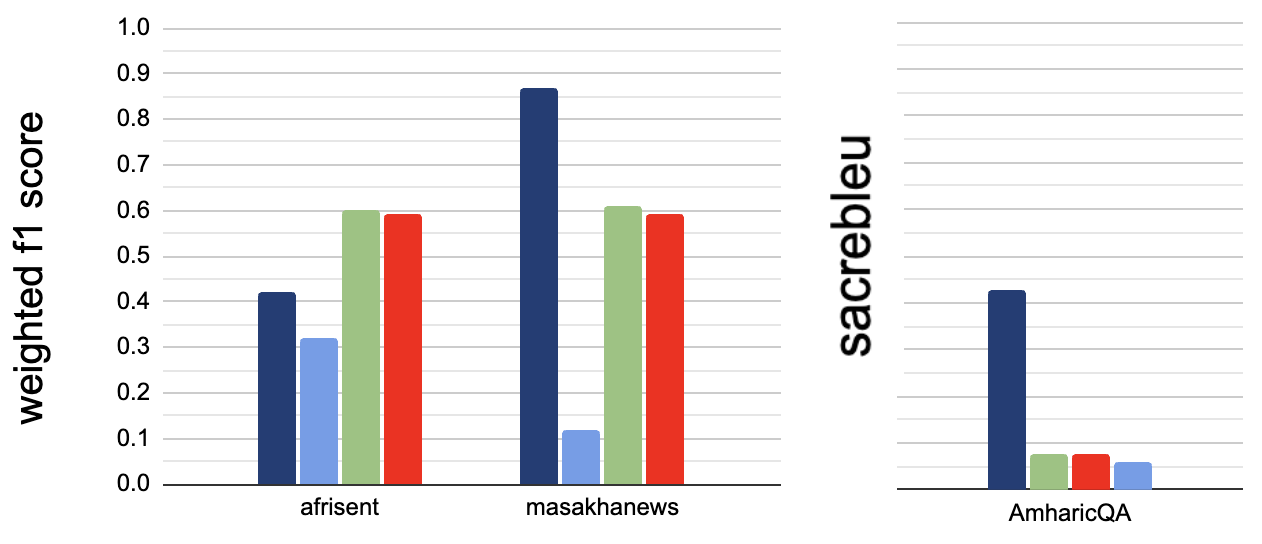} }}%
    \qquad{{\includegraphics[width=7cm]{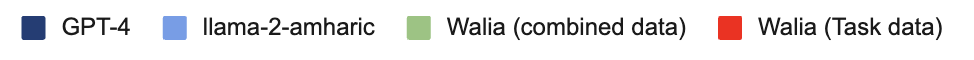} }}
    \caption{Generation scores: weighted f1 scores for AfriSenti and MasakhaNews (left) and SacreBLEU score for  Amharic QA (right)}%
    \label{fig:classification}%
\end{figure}
\begin{figure*}[!ht]%
    \centering
    \centering{{\includegraphics[width=7cm]{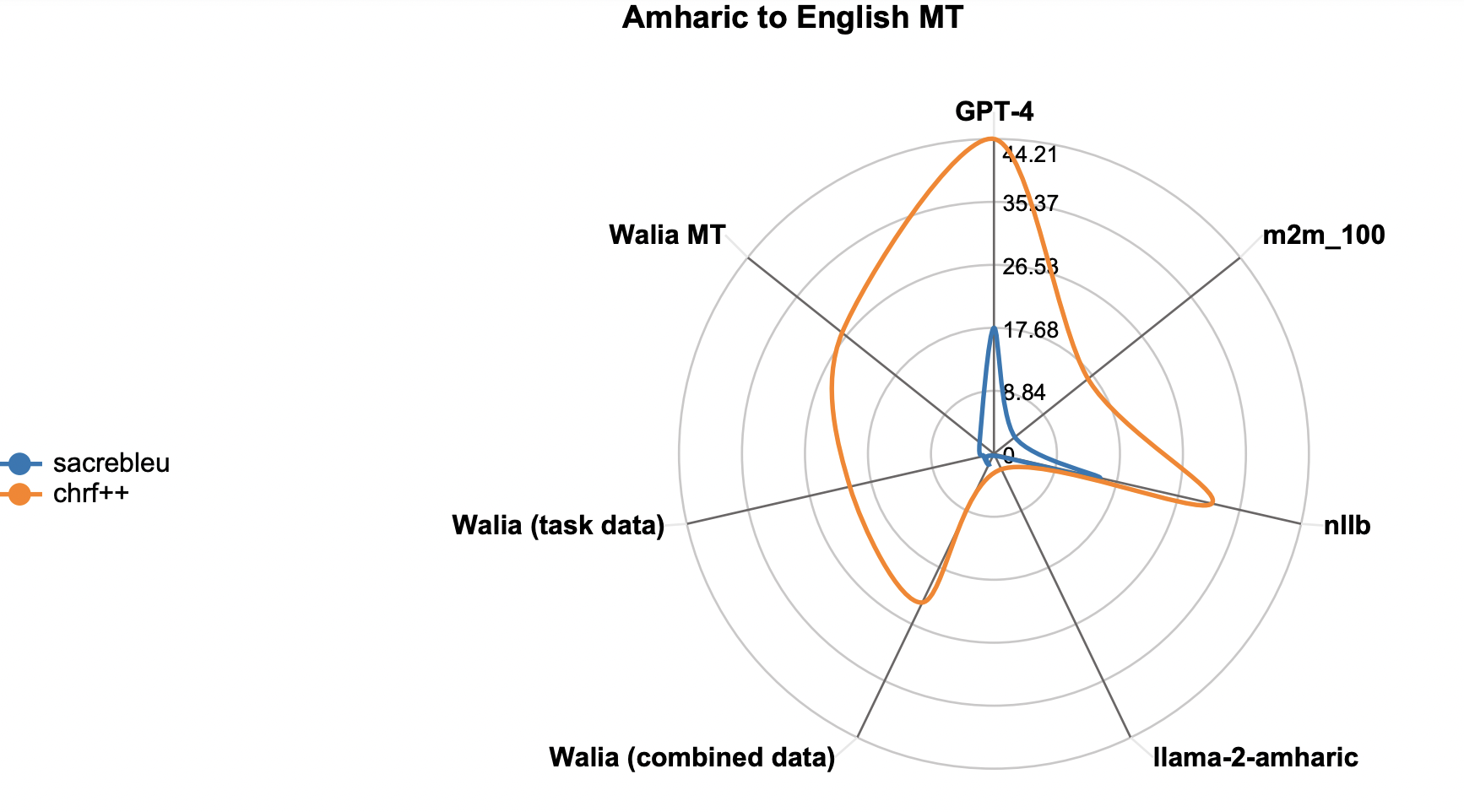} }}%
    \qquad{{\includegraphics[width=7cm]{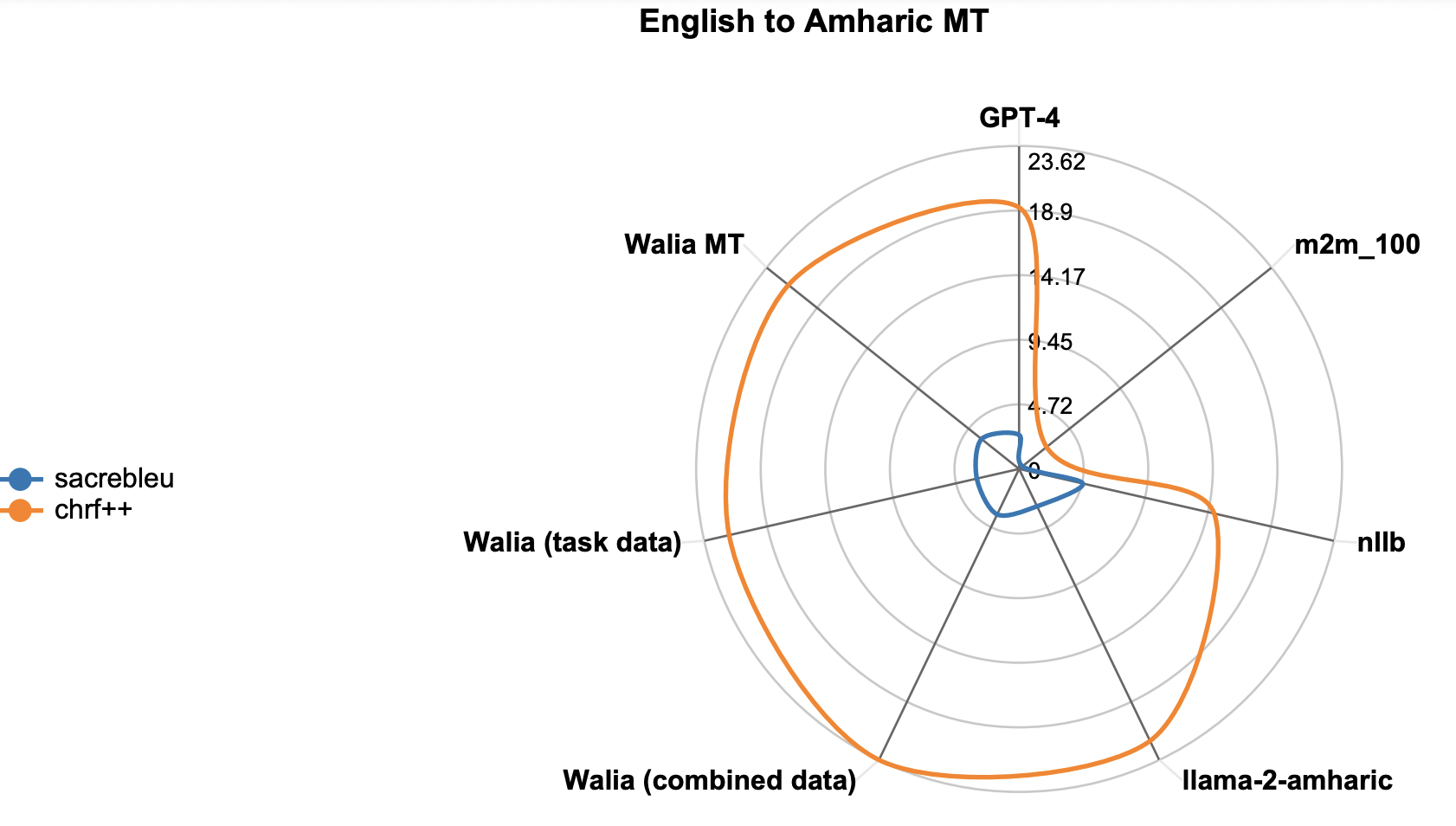} }}%
    \caption{Scores for \textbf{machine translation}. Amharic to English translation scores (Right)  and English to Amharic translation scores( left).  }%
    \label{fig:MT}%
\end{figure*}
\subsection{Prompt based experiments}
\label{sec:prompt}

Throughout our investigation, we observed using only one instruction for a task introduces a high dependency on the prompt, leading to prompt over-fitting. In this case, models fail to do tasks like sentiment classification when presented with different instruction prompts. To deal with this problem, we worked on manually produced templates for each task as shown in Table \ref{tab:newdata}.

Additionally, we experimented with the prompt header part of the dataset shown in \ref{fig:datapipeline}. The prompt header is additional English text stating, \texttt{"Below is an instruction that describes a task. Write a response that appropriately completes the request."}. Introducing this in models like GPT-4  yielded a significant reduction in classifiable outputs, which highlights the effectiveness of incorporating clear English instructions to steer the model toward the desired outcome.

\begin{table}[!ht]
\footnotesize
\centering
\tiny
\begin{tabular}{lrrrrr}
\toprule
\textbf{Tasks}   &  \textbf{GPT-4} & \textbf{\textsc{LLaMA-2}-Am} &\textbf{Walia I} & \textbf{Walia II} \\
\midrule
Text summarization  & 3.34   & 0.62  & 1.13 & 0.80   \\
Text expansion & 3.10 & 3.22 & 2.05  & 2.82   \\
Amharic QA & 28.23 & 2.83 & 5.37 & 6.25 \\
\bottomrule
\end{tabular}
\caption{I = Walia (task data), II = Walia (combined
data). \textbf{ROUGEL} scores for text summarization, Text expansion and Amharic QA.}
\label{tab:rogue}
\end{table}

\section{Results}
Below, we discuss the performance of each model we tested by task type and evaluation strategy. 

\subsection{Classification Results}

For classification tasks, we used two metrics. Our models improve \textsc{Llama-2}-Amharic scores as shown in AfriSenti, MasakhaNews, and QA tasks in Figure \ref{fig:classification}.  The other metrics we reported measure how many times the model returns one of the categories. For the AfriSenti classification task, 759 and 52 out of 1999 test data are not in any of the three classes for \textsc{Llama-2}-Amharic and  GPT-4, respectively. Our models reduce these unusable results and do not produce unusable outputs. For MasakhaNews 248, 136, 106, and 3, results are unusable for \textsc{Llama-2}-Amharic, Walia (task data), Walia (combined data), and GPT-4, respectively. In MasakhaNews case, GPT-4 tends to take the lead in producing reasonable outputs.

\begin{table}[!ht]
\footnotesize
\tiny
\centering
\begin{tabular}{lrrrrr}
\toprule
\textbf{Tasks}   &  \textbf{GPT-4} & \textbf{\textsc{LLaMA-2}-Am}  &\textbf{Walia I} & \textbf{Walia II} \\
\midrule
 Story generation  & 2.93 & 1.00 & \underline{3.60}  & 1.73   \\
 Poem completion  & 2.53 & 1.46 & 1.73  & \underline{2.26}   \\
 Poem generation  & 2.13 & 1.00 & \underline{2.46}  & 2.00   \\
 Religious Lyrics Gen. & 2.86 & 1.46 & \underline{1.60}  & 1.46   \\
  Religious Lyrics Compl. & 3.60 & 1.40 & \underline{2.13}  & 1.93   \\
Non religious Lyrics Gen. & 3.53 & 1.00 & 1.60 & \underline{2.06} \\
\bottomrule
\end{tabular}
\caption{I = Walia (task data), II = Walia (combined data). Average blind \textbf{human evaluation} out of 5, for three people in each task. \textbf{(1)} empty or non Amharic text. \textbf{(2)} not written in task format. \textbf{(3)} written in task format but no consistent idea and spelling errors. \textbf{(4)} looks like that specific generation task but has spelling and grammar errors. \textbf{(5)} this looks like a perfect generation of the task. Underlined text indicates cases where we see improvement compared to \textsc{LLaMA}-2-Amharic.  
}
\label{tab:humaneval}
\end{table}

\subsection{Generation Results}

As explained in Section \ref{Evaluation}, we focus on RogueL metrics for our analysis. Across text summarization and AmharicQA, GPT-4 takes the lead, showing the model's generation ability is very high. We were able to improve the \textsc{LLaMA}-2-Amharic model's ability for this task using our data, as shown in Table \ref{tab:rogue}.

We conducted a human evaluation for the models that do not have fixed gold labels, as shown in Table \ref{tab:humaneval}. Table \ref{tab:humaneval} result shows that the generation ability of L\textsc{Llama-2}-Amharic can be enhanced by adding generation-specific datasets. Walia lacks an understanding of the specific formatting of texts because of the limitations in our pre-processing. However, it shows significant improvement where the \textsc{Llama-2}-Amharic fails to understand the query.

\subsection{Machine Translation (MT)}
For the MT task we evaluated two open-source sequence-to-sequence models (M2M100 \citep{fan2021beyond} and NLLB \citep{nllb2022}),  GPT-4, \textsc{Llama-2}-Amharic, and our models. Figure \ref{fig:MT} shows SacreBLEU and chrF++ results for the above  MT models. As shown in the figure, from MT models, GPT-4 showed better results than the other models when using English as the target language. However, our models showed results comparable to the NLLB and m2m100 models and outperformed the \textsc{Llama-2}-Amharic model for the Amharic-English translation. For the English-Amharic translation, the NLLB model outperformed the others in the SacreBLEU score, while our models showed comparable results and outperformed GPT-4, \textsc{Llama-2}-Amharic and m2m100  models in this translation direction. In our MT evaluation, we noticed irregularities between the results of the two evaluation metrics.  Since SacreBLEU is a word-based metric, the results show that the scores are too low. This shows that using only automatic evaluation metrics makes interpreting and generalizing the results hard. We will add metrics like human evaluation to evaluate MT results in the future.  

\section{Conclusion and Future Works}
In this work, we created Amharic instruction fine-tuning dataset, evaluated the performance of existing and our fine-tuned models in the new dataset, and explored the effect of carefully curated datasets on the models' performance. We observed a possibility of reusing task-specific datasets to improve the generation and task performance of the existing \textsc{Llama-2}-Amharic model. 

Our data generation pipeline that generates instruction datasets from task datasets can be used for the generation of similar datasets for other languages given template instructions. We are working on this kind of dataset for all languages included in MasakaNER \citep{adelani2021masakhaner}, MasakhaNews \citep{adelani2023masakhanews}, AfriSenti \citep{muhammad2023afrisenti} and more to improve multilingual \textsc{Llama} models. We plan to open-source the instruction datasets with the generation code.

 Moving forward, we aim to improve both the quality and volume of the data utilized. Task-specific dataset creations are meant to complement, not replace, language-specific instruction dataset creations, and we plan to work on creating quality instruction datasets in addition to using existing task datasets. We also plan to explore the relevance of using LLMs for evaluation in low-resource languages like Amharic and incorporate LLMs to evaluate our LLMs.

\section{Limitations}

One limitation we observed in our work is the lack of reliable generation metrics for our tasks. The models tend to generate wordy and explained outputs despite our attempts to design the instruction template specifically. As a solution, we used several metrics that can express one task's ability and reported the best-suited one. 

In our current evaluation of all the models, we observed significant limitations while preforming the spell correction and NER tasks. For Amharic spell correction all four generation models, including GPT-4, try to generate other things related to the text, and the word error rate for all of them is close to 99\%.

We have yet to explore the effect of using machine-translated instruction datasets for building language-specific LLMs with regard to introducing cultural bias.

\bibliography{custom}

\onecolumn
\appendix
\section{Experimental details}
\label{sec:appendix}

We adopted our \textsc{LLama-2} instruction tuning experimental code from Chinese-\textsc{LLaMA}-Alpaca\footnote{\url{https://github.com/ymcui/Chinese-LLaMA-Alpaca}} repository by \cite{cui2023efficient}. we performed our instruction frinetuning for three epochs on  4 A100 GPUs using the parameters in Table \ref{tab:param}. In the generation phase, we used parameters in Table \ref{tab:generation} for all models except GPT-4.

\begin{table}[!htb]
    \begin{minipage}{.5\linewidth}
      \centering
        \begin{tabular}{lrrrrr}
            \toprule
            \textbf{Parameter}   &  \textbf{Value} \\
            \midrule
            epoch  & 3  \\ 
            lr  & 1e-4  \\ 
            lora\_rank   & 8   \\ 
            lora\_alpha   & 32  \\
            lora\_dropout  & 0.05  \\
             per\_device\_train\_batch\_size  & 1   \\
             per\_device\_eval\_batch\_size  & 1    \\
             gradient\_accumulation\_steps  & 8    \\
            \bottomrule
            \end{tabular}
            \caption{Training Parameters}
            \label{tab:param}
    \end{minipage}%
    \begin{minipage}{.5\linewidth}
      \centering
        \begin{tabular}{ll}
            \toprule
            \textbf{Parameter} & \textbf{Value} \\
            \midrule
            seed & 42 \\
            do\_sample & True \\
            min\_length & None \\
            top\_p & 1.0 \\
            temperature & 1.0 \\
            top\_k & 5 \\
            repetition\_penalty & 5.0 \\
            length\_penalty & 1 \\
            \bottomrule
            \end{tabular}
            \caption{Configuration settings for token generation.}
            \label{tab:generation}
    \end{minipage} 
\end{table}

\section{Dataset details}
\label{sec:appendix2}
\begin{figure*}[!ht]
    \centering
\includegraphics[width=16cm]{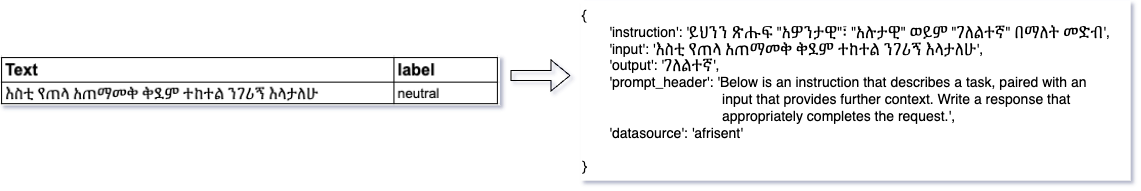}
    \caption{Example data output from our dataset creation pipeline. \ref{tab:newdata}}
    \label{fig:example-pipeline2}
\end{figure*}

Figure \ref{fig:example-pipeline2} shows how we repurposed existing sentiment analysis data to convert it into an instruction dataset. We utilized a task template selected at random from our collection. The number of templates collected for each task is shown in the table. The reason for keeping the prompt header in the English language is discussed in Section \ref{sec:prompt}.

\begin{figure*}[!ht]
    \centering
\includegraphics[width=12cm]{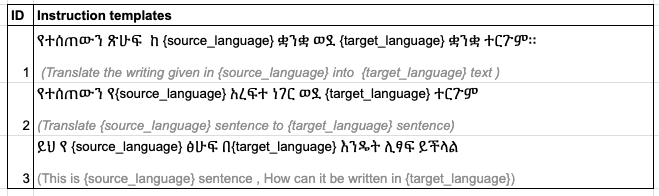}
    \caption{Example templates for machine translation task with English translation. By using random instructions for tasks, we ensure that the model does not fit the specific instructions for tasks. }
    \label{fig:datapipeline2}
\end{figure*}

To avoid instruction overfitting, as discussed in Section \ref{sec:prompt}, we collected a variety of instructions, like the example shown in Figure \ref{fig:datapipeline2}. The example displays different instructions that can be paired with a machine translation dataset to create a machine translation instruction dataset. This step is repeated for all tasks, including tasks like poem generation, which we created by collecting from different websites.

\begin{figure*}[!ht]
    \centering
\includegraphics[width=7cm]{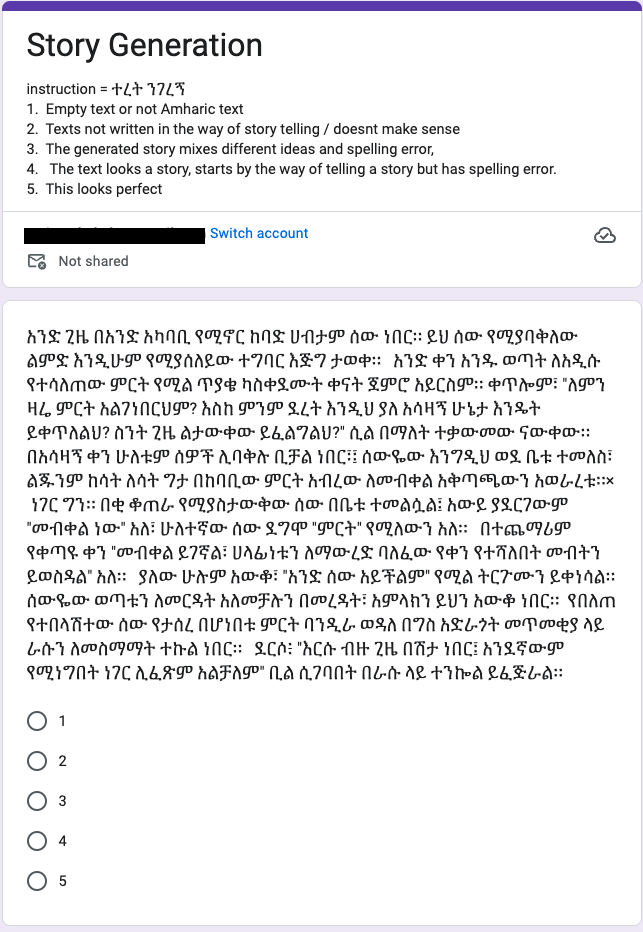}
    \caption{Form used for human annotation with labeling instruction. We see in the figure how one question of the sample story generation task is being validated.}
    \label{fig:humanannotation}
\end{figure*}

Due to the difficulty we faced in evaluating generation tasks, we employed human evaluation. Figure \ref{fig:humanannotation} shows how evaluators scored each generation output for the case of the story generation task.

\section{Result details}
In Table \ref{tab:rogue2}, we present detailed scores for text summarizing, text expansion, and Amharic QA. The choice of the right parameter needs further exploration but RogueL is a good metrics to show improvement because it doesn't depend on specific n-gram similarities.

\begin{figure*}[!ht]
    \centering
\includegraphics[width=10cm]{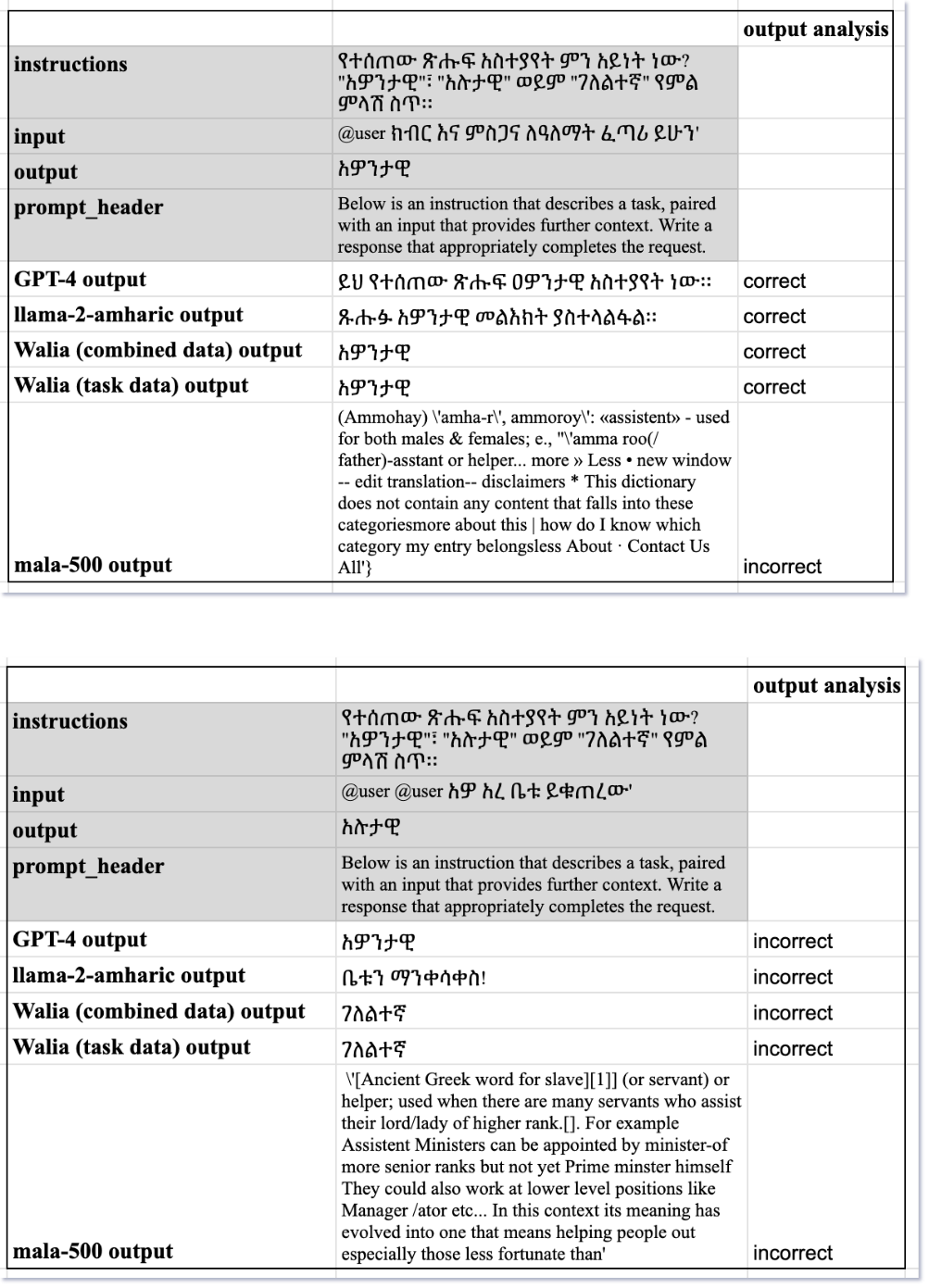}
    \caption{Example of analysis we did on the model outputs. }
    \label{fig:analysis}
\end{figure*}

In Figure \ref{fig:analysis} we demonstrate that even if the model's output is not an exact match, we have created a pipeline to verify and identify the correct output from the generated sequence. We limit our-self to GPT-4 as an external model because it's not explored in other well-known models. Additionally, we reveal that the mala-500 \cite{lin2024mala500} model produces unrelated outputs, which merits deeper examination.

\begin{table*}[!ht]
\footnotesize
\centering
\scriptsize
\begin{tabular}{lrrrrr}
\toprule
\textbf{Tasks}  &  \textbf{GPT-4} & \textbf{LLaMA-2-Amharic} &\textbf{Walia (Task data) } & \textbf{Walia (combined data)} \\
\midrule
Text summarization  & 3.41/0.11/3.34   & 0.61/0.00/0.62  & 1.12/0.00/1.13 & 0.78/0.00/0.80    \\
Text expansion & 3.11/0.11/3.10 & 3.35/0.02/3.22 & 2.14/0.02/2.05  & 2.89/0.10/2.82   \\
Amharic QA & 28.22/8.00/28.23 & 2.83/0.66/2.83 & 5.36/0.67/5.37 & 6.34/1.56/6.25 \\
\bottomrule
\end{tabular}
\caption{\textbf{Rogue1/Rogue2/RogueL} scores for text summarization, Text expansion and AmharicQA}
\label{tab:rogue2}
\end{table*}
\end{document}